\documentclass[sigconf]{acmart}

\usepackage{booktabs} 
\renewcommand\footnotetextcopyrightpermission[1]{}
\begin{document}
\title{Network Modeling and Pathway Inference from Incomplete Data ("PathInf")}
\author{Xiang Li}
\authornote{Equal contributions}
\orcid{0000-0002-9851-6376}
\affiliation{
  \institution{MGH/BWH Center for Clinical Data Science}
  \city{Boston}
  \state{MA}
}
\email{xli60@mgh.harvard.edu}

\author{Qitian Chen}
\authornotemark[1]
\affiliation{
  \institution{Center for Data Science, Peking University}
  \city{Beijing}
  \country{China}
}
\email{chenqitian.sweet@163.com}

\author{Xing Wang}
\authornotemark[1]
\affiliation{
  \institution{Peking University Cancer Hospital \& Institute}
  \city{Beijing}
  \country{China}
} 
\email{15901013210@126.com}

\author{Ning Guo}
\affiliation{
  \institution{Department of Radiology, Massachusetts General Hospital}
  \city{Boston}
  \state{MA}
}
\email{Guo.Ning@mgh.harvard.edu}

\author{Nan Wu}
\authornote{Joint corresponding authors}
\affiliation{
 \institution{Peking University Cancer Hospital \& Institute}
 \city{Beijing}
  \country{China}
}
\email{nanwu@bjmu.edu.cn}

\author{Quanzheng Li}
\authornotemark[2]
\affiliation{
  \institution{MGH/BWH Center for Clinical Data Science}
  \city{Boston}
  \state{MA}
}
\email{li.quanzheng@mgh.harvard.edu}

\begin{abstract}
In this work, we developed a network inference method from incomplete data ("PathInf") , as massive and non-uniformly distributed missing values is a common challenge in practical problems. PathInf is a two-stages inference model. In the first stage, it applies a data summarization model based on maximum likelihood to deal with the massive distributed missing values by transforming the observation-wise items in the data into state matrix. In the second stage, transition pattern (i.e. pathway) among variables is inferred as a graph inference problem solved by greedy algorithm with constraints. The proposed method was validated and compared with the state-of-art Bayesian network method on the simulation data, and shown consistently superior performance. By applying the PathInf on the lymph vascular metastasis data, we obtained the holistic pathways of the lymph node metastasis with novel discoveries on the jumping metastasis among nodes that are physically apart. The discovery indicates the possible presence of sentinel node groups in the lung lymph nodes which have been previously speculated yet never found. The pathway map can also improve the current dissection examination protocol for better individualized treatment planning, for higher diagnostic accuracy and reducing the patients trauma.
\end{abstract}
\maketitle
\section{Introduction}
\begin{figure}
\centering
\includegraphics[width =\columnwidth]{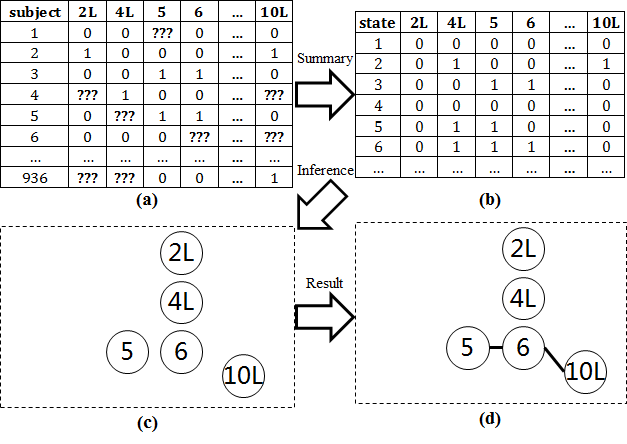}
\caption{Illustration of the algorithmic pipeline. The binary metastasis matrix with missing values (a) is used to estimate the data summarization state matrix (b). Starting from an empty graph (c), the final network (d) is inferred from the state matrix. Due to the space constraint only a portion of the nodes are shown in the figure.}
\label{fig:Fig1}
\end{figure}

Bayesian Network modeling has been a useful tool to infer causal relationships in various applications \cite{doi:10.1093/restud/rdr004,doi:10.1002/hbm.22404,AGUILERA20111376,Jiang2011,SUN2012987}. In recent years various algorithms have been developed based on the concept of Bayesian Network. To name a few, works in \cite{RN969} used equivalence class to find a Bayesian Network which can be a perfect map of given distribution when the sample size is large enough. Works in \cite{RN968} applies an information-theoretic approach to efficiently learn structure from data. Both of these works are based on the conditional independence. Traditionally, speed and accuracy of structure learning algorithms are limited for more complex data. Models proposed in \cite{RN970} and \cite{RN975} tried to lower the computational cost in order to tackle more complex cases, which all require complete data observation. The Greedy Equivalent Search (GES) method proposed in \cite{RN969} quickly become the standard in network learning due to its accuracy and capability in handling large dataset.
One of the major challenge in network inference is the uncertainty and missing values in the data (i.e. incomplete data). To deal with such challenges, \cite{Friedman:1997:LBN:645526.657145} tries to find the optimal Bayesian Network structure with model selection within the process of Expectation-Maximization (EM). This work is among the first attempts to utilize EM algorithm during the structure learning. Later time, \cite{RN972} proposes a combination of GES and EM algorithm to learn the network structure form incomplete data. To the best of our knowledge, the GES-EM method introduced in \cite{RN972} is the best solution towards the task of inferring a graph representation of network from incomplete data. However, for applications in real-world settings, these methods suffer from the following limitations:
\begin{enumerate}
  \item They all need relatively larger sample size in order to perform feasible search over the optimized structures.
  \item Most of the algorithms cannot correctly handle missing values.
  \item The Directed Acyclic Graph (DAG) assumption underlying all Bayesian networks might not be valid in real practice, as circles and/or two-way connections among variables are common.
\end{enumerate}
In response to the above challenges for current network modeling methods, we develop a two-stages Bayesian Network inference model ("PathInf"). In the first stage, maximum likelihood based data summarization model is applied to overcome the presence of massive, non-uniformly distributed missing values in the data. In the second stage, after obtaining the state matrix through the data summarization model, the transition patterns among variables (i.e. pathway) are inferred as a graph inference problem solved by greedy algorithm with constraints. Experiment results on simulation data shows that PathInf can overcome these challenges and achieve better accuracy in recovering the true networks. 
In addition, we use PathInf to analyze the lung cancer metastasis data, for the inference of transition pathways among lymph vascular nodes. While there have been various literature on lymphatic spreading \cite{RN961, RN962, RN963, RN964, RN965}, most of the previous conclusions were made by clinical physicians and focused on specific empirical transition paths, rather than a holistic map depicting all possible metastasis pathways simultaneously. From a precision medicine perspective, investigation and modeling of lymphatic metastasis pathway would be vital for its predicting value on the different roles of lymph nodes during the cancer spreading, and its potential to provide individualized surgical planning strategy. By applying PathInf on the lymphatic spreading data from 936 patients diagnosed with non-small-cell lung cancer at various cancer stages, we obtain the map of possible lung cancer metastasis pathways. Preliminary conclusion from the map shows that the proposed data-driven PathInf method can obtain similar lymphatic metastasis pathways as found in the previous work based on clinical and anatomical studies. In addition, novel and consistent metastasis interrelationship unknown to the physicians have been found in the network, showing potential jumping metastasis among lymph nodes. The result also indicates the need for an improved lung cancer staging and diagnosis criteria. An illustrative pipeline showing the input (binary metastasis data), data summarization model and graph inference model is visualized in ~\ref{fig:Fig1}.
\section{MAXIMUM LIKELIHOOD ESTIMATION ON DATA SUMMARIZATION WITH MISSING VALUES}
Given a binary (e.g. cancer dissection examination result on lymph nodes, which is positive/negative) data matrix $X$ with $n$ variables, all the possible combinations of the binary values in these $n$ variables can be summarized into the state representation $\boldsymbol{S}=\{S_1,\ldots,S_{2^n}\}$. States $\boldsymbol{S^X}$ that are conformed to $X$ (i.e. capable of representing all samples in $X$ with minimum number of states) is a subset of $\boldsymbol{S}$ which can be estimated. In practice, the true $X$ is often hidden and its observations are incomplete, resulting in the observation matrix $O$ with missing values. Due to the presence of missing values, states $\boldsymbol{S}^O$ that are conformed to $O$ are ambiguous. With prior information on the proportion of missing values, it is possible to estimate $\boldsymbol{S}^O$ so that it can optimally describe the observations, which is our main rationale for the data summarization model.
Most of the previous work on learning network from incomplete data adopts the missing at random assumption \cite{RN972}. However, the distribution of missing values in practice commonly follows a biased distribution caused by latent variables. For example, in lymphatic spreading data, dissection and examination of the lymph nodes depends on the surgical conditions and physicians decision, which can follow a specific pattern depending on severity of the cancer. For example, cancer with tumor nodules smaller in size can result in a sparsely-examined lymph node data (i.e. more missing values). To overcome this challenge of massive and non-uniformly distributed missing values, we estimate of the state representation $\boldsymbol{S^O}$ using maximum likelihood method with two priors: the probability of a negative value being missed and the probability of a positive value being checked. Specifically, assuming that the presence of missing values in each sample of $X$ are independent, based on the total probability theorem, the probability of observing $O_i$ can be defined as:
\begin{equation}\label {Eq:1}
\begin{aligned}
&P(O_i)=\sum_{j}^{2^n}{P(O_i|S_j)P(S_j)} \\
&P(O_i|S_j)=
\begin{cases}
0.5^n(\frac{p}{1-p})^{x_i}(\frac{1-p}{0.5})^{y_j}, & \text{if } O_i\in S_j \\
0 & \text{otherwise, }
\end{cases}
\end{aligned}
\end{equation}
where $x_i$ is the count of observed positive values in $O_i$, $y_j$ is the count of positive values in $S_j$. Notation $O_i\in S_j$ indicates that assuming the missing values can be both positive or negative, state $S_j$ is conformed with sample $O_i$. The probability of a missed negative value is set to the constant of 0.5, as we are not focused on how the negative samples are handled. The probability of a missed positive value, which is far more important, is set to $p$ which ranges from 0.1 to 0.4 in this work, to reflect the fact that positive lymph nodes are more likely to be dissected and examined. 
Based on Eq.\ref{Eq:1}, given the observed dataset $\boldsymbol{O}=\{O_1,\ldots,O_m\}$, the estimation of its state representation $\boldsymbol{S^O}$ can then be formulated as a constrained optimization problem in order to get the maximum likelihood estimation for its probability distribution: 
\begin{equation}\label {Eq:2}
\begin{aligned}
&\arg\min\limits_{P(S_j)}(-\sum_{i}^{m}{\ln P(O_i)}),\\
&\text{s.t. } \sum_{j}^{2^n}{P(S_j)}=1 \text{ and } 0\leq P(S_j)\leq 1, \forall j.
\end{aligned}
\end{equation}
Results from the estimation are the probability of each state representation in $\boldsymbol{S}$ based on data $\boldsymbol{O}$. As in the real-world scenarios the network structure tends to be sparse, states in $\boldsymbol{S}$ with non-zero probability in the estimation result will only be a small portion of the whole combination set. Denoting $Ind=\{i_1,\ldots,i_q\}$ as indices of all non-zero elements in the estimated probability $\boldsymbol{P}$, the state representation $\boldsymbol{S^O}$ of $\boldsymbol{O}$ can then be represented as $\{S_{i_1},\ldots,S_{i_q}\}$. As both the objective function feasible set are convex in Eq.\ref{Eq:2}, this optimization problem is convex. Proximal gradient descent is used to solve this optimization problem, as formulated below:\\
\begin{table}
\centering
\renewcommand{\arraystretch}{1.2}
\begin{tabular}{|c|c|}
\hline
\multicolumn{2}{|c|}{Notations} \\ \hline
$\boldsymbol{P}$ & Random vector $(P(S_1 ),\ldots,P(S_{2^n}))$ \\ \hline
$F(\boldsymbol{P})$ & Objective function $-\sum_{i}\ln P(O_i)$ \\ \hline
$\bigtriangledown F$ & Gradient of $F(\boldsymbol{P})$ \\ \hline
$C$ & Feasible set $C=\{\boldsymbol{P}|\textbf{1}^T \boldsymbol{P}=1, \textbf{0} \leq \boldsymbol{P} \leq \textbf{1}\}$ \\ \hline
$\textrm{I}_C(\boldsymbol{P})$ & Indicator function, function value equals to $0$ \\ & if $\boldsymbol{P}$ is in $C$, otherwise equals to the infinity. \\ \hline
$prox_f(x)$ & Proximity operator, \\ & $prox_f(x)=\arg\min_u \frac{1}{2}\left \| x-u \right \|_2^2 +f(u)$ \\ \hline
\end{tabular}
\end{table}
The optimization problem can be solved by: Initialization by randomly choosing $\boldsymbol{P}_0$ from $C$, followed by iteratively updating $\boldsymbol{P}$ use the rule $\boldsymbol{P}_k+1=prox_{I_C}(\boldsymbol{P}_k-\bigtriangledown F(\boldsymbol{P}_k))$ until convergence.\\ 
\begin{figure}
\centering
\includegraphics[width =\columnwidth]{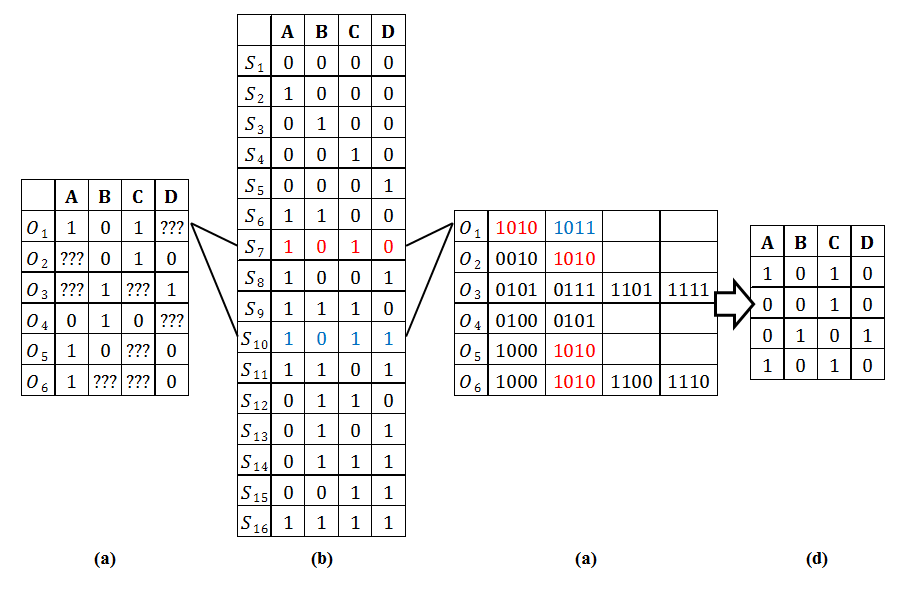}
\caption{Running example of the data summarization modeling. (a) Model input. (b) All possible states. (c) States that are conformed to the input. (d) Final estimation result showing the state with probability greater than zero. Probability of all other states are zero.}
\label{fig:Fig2}
\end{figure}
A running example of the data summarization model is shown in ~\ref{fig:Fig2}. In the figure, given an input matrix (a) with 4 variables $(A, B, C, D)$ and 6 samples $(O_1 \sim O_6)$ with missing values, there are totally 16 different combination of the binary values (b). For each sample with missing value, multiple states can conform to it (b-c, where conformed states to sample $O_1$ are colored in red and blue). Maximum likelihood method is then used to estimate the probability of the state representations (c), resulting in the summarization matrix (d). For example, the posterior probability of state "$1010$" (marked by red) is larger than "$1011$" (marked by blue). Thus during the estimation, state "$1010$" will be kept while "$1011$" will be discarded, because it is more optimized to set the probability of state "$1011$" to 0 in order to increase the probability of state "$1010$" than any other configurations. It can be seen that the maximum likelihood estimation can obtain a sparse and accurate representation of the input data, which is exactly what we need for this study.
\section{GRAPH INFERENCE FROM SUMMARIZATION MATRIX}
In this section, we propose a greedy method to infer the network from the summarized state matrix with the constraint of minimizing the number of edges. Based on the results from the previous section, the state of the input observation has been fully described by $S^O$ with a limited number of combinations. Given an undirected graph $G=(V, E)$, $V$ representing the vertex set and $E$ representing the edge set of $G$, for any state $S$ representing a specific combination of the n binary values, we define:
\begin{itemize}
	\item $V_S$ is the vertices set of positive values in $S$.
	\item $E_S$ is the set consisting of all edges in the complete graph on $V_S$. 
	\item Connected Vertices: two vertices with a path between them.
	\item Connected Graph: graph in which every pair of vertices is connected vertices.
	\item Connected Component: two vertices are in the same connected component if and only if they are connected. \end{itemize}
The problem of network inference is then proposed as follows:
\begin{itemize}
	\item Given state $\boldsymbol{S^O}=\{S_{i_1},\ldots,S_{i_q}\}$, identifying a graph G with minimum edges which satisfies that: $\forall S \in \boldsymbol{S^O}$, the subgraph $G_j=(V_j, E \cap E_j)$ is connected.
\end{itemize}
The rationale for this inference process is that resulting network is supposed to explain all the co-existence of the positive values in each state $S$ from the given set $\boldsymbol{S^O}$. At the same time the total number of links in the network should be minimized, as otherwise a fully connected graph with $n$ vertices can also satisfy the connected subgraph condition. Based on the proposal above, the network is inferred by the following steps. It should be noticed that the inference procedure does not take the acyclic assumption for the edges, as the definition of connected vertices and connected graph only consider connectivity without direction. So the model should perform the same for cyclic or acyclic graphs. 
\begin{enumerate}
  \item The algorithm begins with an empty graph $G=(V, E)$ with $n$ vertices and no edges.
  \item For $\forall S \in \boldsymbol{S^O}$, if $\boldsymbol{V_S}$ contains only two vertices, add an edge between these two vertices. 
  \item For any two disconnected vertices $x$ and $y$ in $G$ and any state $S \in \boldsymbol{S^O}$, if $x, y \in \boldsymbol{V_S}$, we define the score $w_S(x,y)$ for a potential edge to be added between $x$ and $y$:
\begin{equation*}
w_S(x,y)=\frac{P(S)}{\# \text{connected components in }G_S},
\end{equation*}
		where $P(S)$ is the probability of the given state as estimated in the previous section.
  \item Calculate $W(x,y)=\sum\limits_{\boldsymbol S^O} w_s(x,y)$, that is, score of an edge between vertices $x$ and $y$ is the sum of scores calculated over all feasible states.
  \item Find vertices pair $(\hat{x},\hat{y})$ with the highest score $W(\hat{x},\hat{y})$ through exhaustion. Then add an edge between each pair in $G$.
  \item Iteratively execute step 3-5 to update $G$, until all $G_S$ are connected graph. 
\end{enumerate}
\section{EXPERIMENTAL RESULTS ON SIMULATION AND APPLICATION DATA}
\subsection{Model validation and performance evaluation with simulation data}
\begin{figure}
\centering
\includegraphics[width =\columnwidth]{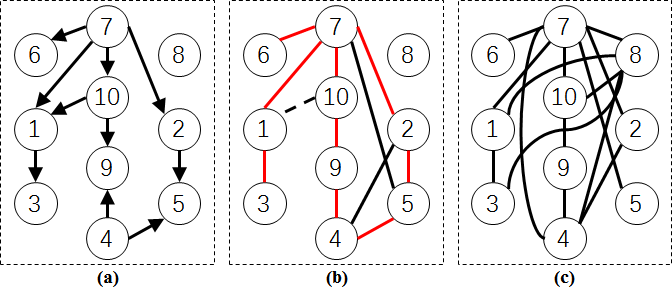}
\caption{(a) Ground truth graph used to generate simulation data observations. (b) Graph inferred by the proposed PathInf method. (c) Graph inferred by GES.}
\label{fig:Fig3}
\end{figure}

\begin{figure}
\centering
\includegraphics[width =\columnwidth]{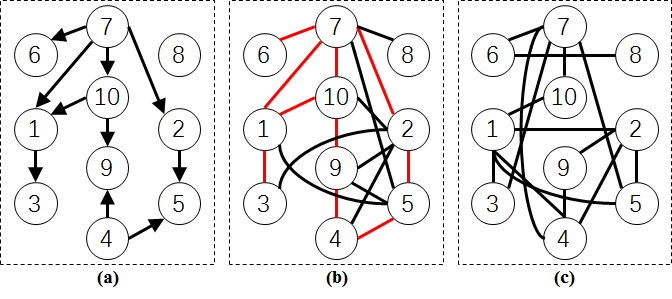}
\caption{(a) Ground truth graph used to generate simulation data observations with missing values. (b) Graph inferred by PathInf. (c) Graph inferred by GES.}
\label{fig:Fig4}
\end{figure}

In order to test the validity of our algorithm, we generate random directed acyclic graphs (DAGs) representing the true pathways among variables, and use discrete time model simulating the spreading process for the observation data. For each given DAG, nodes with indegree of zero (i.e. source nodes) are selected as the starting nodes for the spreading, as they cannot be the target from other nodes. As the graph is acyclic, at least one node is the source node which can serve as starting node. The following steps are then applied to generate observation data:
\begin{enumerate}
  \item Generate a random DAG with fixed number of nodes (10) and various predefined number of edges (we tested 10, 15, 20, 25 in this study).
  \item  Uniform random weights are assigned to the edges. Weight of edge connecting node $i$ to node $j$ is denoted as $w(i,j)$. $w(i,j)$ is zero if there is no edge. 
  \item Identify the source node(s) in DAG and select them as starting node(s) for the spreading.  
  \item Apply Markov Chain Monte Carlo (MCMC)-based sampling method to generate each observation sample:
  \begin{enumerate}
    \item Set the number of total transition steps as a random number following Poisson distribution. Larger total transition steps result in more positive nodes in the generated sample.
    \item If there are more than one source nodes in the graph, randomly select one as the starting node.
    \item In each transfer step, denote the positive nodes as $A=\{a_0,\ldots,a_m\}$, and the nodes that are connected from $A$ as set $B=\{b_1,\ldots,b_n\}$. If $B$ is nonempty, the probability of node $b_i$ to become positive (i.e. spreading) is $\sum_{j}^{m} w(a_j,b_i) / \sum_{k}^{n}\sum_{j}^{m} w(a_j,b_k)$. 
    \item Transition process stops when B is empty or reaching the total number of transition.
    \item After the sample is generated, missing values are added to replace original values. The probability for a positive node to become missing is $p$ ($p<0.5$, we tested 0.1 to 0.4). The probability for a negative node to become missing is 0.5.
  \end{enumerate}
  \item For each simulation dataset, repeat Step 4 to generate 1000 samples with the same DAG but different transition pathway.
\end{enumerate}
A sample result showing the ground truth network, along with the network inferred by our method (PathInf) and Greedy Equivalent Search (GES) method without missing values is shown in ~\ref{fig:Fig3}. It can be seen that when the data quality is high, PathInf could infer most of the edges from the ground truth graph (colored as red), with both very few false positive (colored as black) and false negative (colored as dashed) edges. On the contrary, GES overestimates the graph, resulting in large number of false positive edges. Given the fact that the input data is free of missing values in this experiment, the lowered performance of GES might be caused by the relatively smaller sample size (1000 samples to infer edges among 10 variables). However, in practice especially for medical data, large sample size is often difficult or even impossible to meet. The high accuracy of PathInf in the experiment shows it can overcome the challenge in sample size, to certain extent. Further, the computation time for PathInf and GES is almost the same, showing its efficiency in making the reference. 
In ~\ref{fig:Fig4} we show the results of PathInf and GES by applying them on the data with large number of missing values ($p=0.4$, indicating 40\% of the positive values are missing and 50\% of the negative values are missing). As the method introduced in \cite{RN972} for analyzing data with missing values using GES plus Expectation-Maximization (EM) algorithm failed to converge on the simulation data, we simply replaced the missing values by zero entries and used original GES method for the inference. The results from GES deviated much from the ground truth, while PathInf can recover the network structure with no false negatives. The added false positive edges of the PathInf results on incomplete data are reasonable, as missing values in the observation expand the state representation matrix SO, which supports more possible edges. 
Finally, we run the simulation and the network inference with different number of edges in the ground truth graph and different values of p (proportion of missing values). The performance statistics are summarized in Table 1, showing that in most of the experiments PathInf can outperform GES especially with regarding to the false negative rate, which is more important for medical applications as it is vital to ensure all possible pathways have been discovered from the data in order to establish a criteria/protocol. It should be noted that as the value of p increases, the false negative rate decreases for PathInf, which can also be observed in ~\ref{fig:Fig3} and ~\ref{fig:Fig4}. This is because a larger $p$ will result in denser inference by PathInf as discussed above, thus more true edges will also be recovered as a result.  
\begin{table}
\caption {Performance of PathInf and GES on simulation data, evaluated by false positive and false negative rate.} 
\centering
\renewcommand{\arraystretch}{1.2}
\begin{tabular}{|c|c|c|c|c|c|c|c|c|}
\hline
 & \multicolumn{4}{c|}{\textbf{PathInf: False Positive}} & \multicolumn{4}{c|}{\textbf{GES: False Positive}} \\ \hline
\begin{tabular}[c]{@{}c@{}}$p$/\\ edges\end{tabular} & 0.1 & 0.2 & 0.3 & 0.4 & 0.1 & 0.2 & 0.3 & 0.4 \\ \hline
10 & 94\% & 96\% & 99\% & 102\% & 131\% & 145\% & 156\% & 150\% \\ \hline
15 & 67\% & 78\% & 76\% & 79\% & 66\% & 77\% & 75\% & 75\% \\ \hline
20 & 52\% & 57\% & 60\% & 60\% & 39\% & 44\% & 41\% & 44\% \\ \hline
25 & 38\% & 41\% & 44\% & 45\% & 25\% & 25\% & 28\% & 36\% \\ \hline
 & \multicolumn{4}{c|}{\textbf{PathInf: False Positive}} & \multicolumn{4}{c|}{\textbf{GES: False Positive}} \\ \hline
10 & 22\% & 19\% & 17\% & 15\% & 36\% & 40\% & 41\% & 43\% \\ \hline
15 & 29\% & 25\% & 23\% & 22\% & 45\% & 46\% & 48\% & 51\% \\ \hline
20 & 30\% & 26\% & 24\% & 23\% & 46\% & 50\% & 52\% & 55\% \\ \hline
25 & 31\% & 27\% & 23\% & 22\% & 48\% & 53\% & 54\% & 58\% \\ \hline
\end{tabular}
\end{table}
\subsection{Metastasis pathway inferred from lung cancer population study}
\begin{figure}
\centering
\includegraphics[width =\columnwidth]{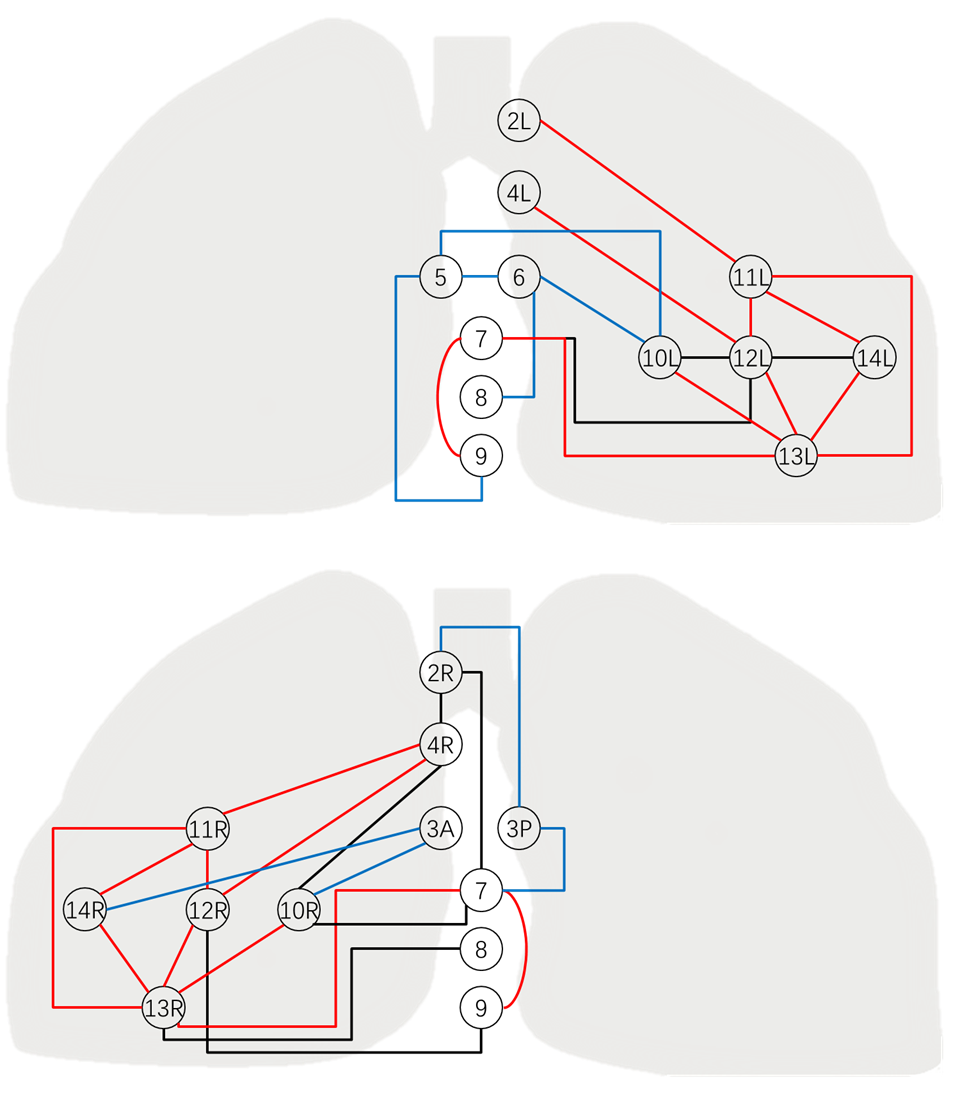}
\caption{Top: pathway graph inferred from the summarization of the left lung. Bottom: pathway graph inferred from the summarization of the right lung. Lymph node sites are shown as round circles in the figure according to their anatomical locations. The inferred transition paths are shown as edges connecting them. Edges colored in red are those consistently found in the left and right lung (i.e. connecting the same nodes). Edges colored in blue are connecting to the unique nodes at each side thus cannot be compared (e.g. edges connecting nodes 5 and 6 can only be found in the left lung).}
\label{fig:Fig5}
\end{figure}

\begin{figure}
\centering
\includegraphics[width =\columnwidth]{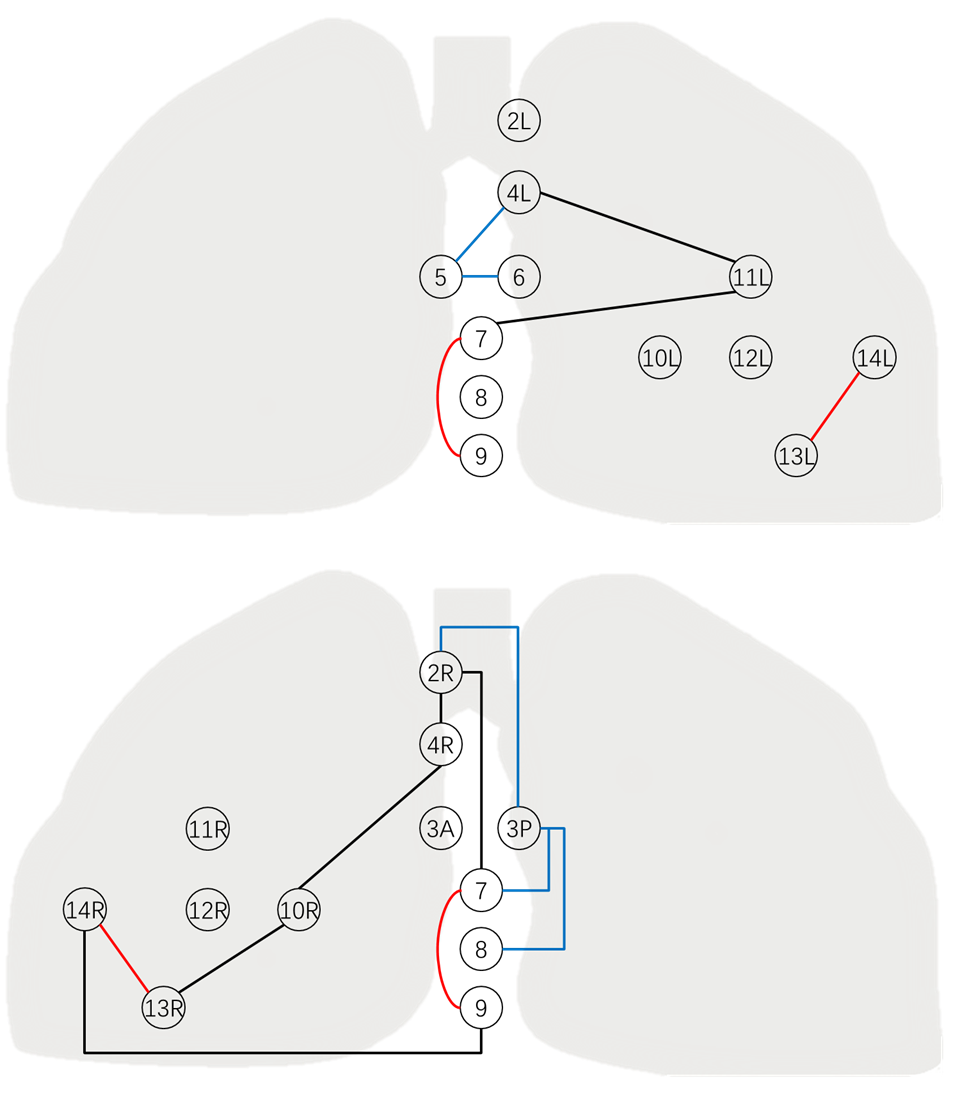}
\caption{Top: pathway graph inferred from the binary observation matrix of left lung by GES. Bottom: pathway graph inferred from right lung. Lymph node sites are shown as round circles in the figure according to their anatomical locations. The inferred transition paths are shown as edges connecting them. Edges colored in red are those consistently found in the left and right lung (i.e. connecting the same nodes). Edges colored in blue are connecting to the unique nodes at each side thus cannot be compared (e.g. edges connecting nodes 5 and 6 can only be found in the left lung).}
\label{fig:Fig6}
\end{figure}

For the data collection project, we perform the complete lymph node examination of a total of 936 subjects diagnosed with lung cancer. Lymph nodes sampling or dissection are performed according to NCCN guidelines, and segmental station 13 and subsegmental station 14 are retrieved routinely. Samples of tissue and lymph nodes are sent for routine pathologic analysis with paraffin-embedded blocks. Lymph nodes area bi-valved along their longitudinal axis and totally submitted for microscopic evaluation. Small nodes (0.4 cm) area submitted without bi-valving. A single hematoxylineeosinestained slide is prepared from each block.
Binary information of the metastasis (i.e. whether cancer cells are observed at the given node) is then recorded at 21 lymph node sites within the chest, forming a 21$\times$936 matrix characterizing the lung cancer metastasis of the population. These sites are located at left/right lung lobes, as well as mediastinum (membranous partition between the lungs). The anatomic landmarks delineating each node are listed in Table 2, derived from the regional lymph node classification system \cite{RN960}. 
\begin{table}
\caption {Name and location of the lymph node sites used in this study, separated into the left and right groups.} 
\centering
\renewcommand{\arraystretch}{1.2}
\begin{tabular}{|c|c|c|c|}
\hline
\multicolumn{2}{|c|}{\textbf{Left Lung}} & \multicolumn{2}{c|}{\textbf{Right Lung}} \\ \hline
\textbf{Node} & \textbf{Site} & \textbf{Node} & \textbf{Site} \\ \hline
2L & \begin{tabular}[c]{@{}c@{}}Upper\\   Paratracheal (left)\end{tabular} & 2R & \begin{tabular}[c]{@{}c@{}}Upper\\   Paratracheal (right)\end{tabular} \\ \hline
4L & \begin{tabular}[c]{@{}c@{}}Lower\\   Paratracheal (left)\end{tabular} & 4R & \begin{tabular}[c]{@{}c@{}}Lower\\   Paratracheal (right)\end{tabular} \\ \hline
5 & Subaortic & 3P & Retrotracheal \\ \hline
6 & Para-aortic & 3A & Pretracheal \\ \hline
10L & Hilar (Left) & 10R & \begin{tabular}[c]{@{}c@{}}Hilar (right)\end{tabular} \\ \hline
11L & Interlobar (left) & 11R & \begin{tabular}[c]{@{}c@{}}Interlobar (right)\end{tabular} \\ \hline
12L & \begin{tabular}[c]{@{}c@{}}Lobar (left)\end{tabular} & 12R & \begin{tabular}[c]{@{}c@{}}Lobar (right)\end{tabular} \\ \hline
13L & \begin{tabular}[c]{@{}c@{}}Segmental (left)\end{tabular} & 13R & \begin{tabular}[c]{@{}c@{}}Segmental (right)\end{tabular} \\ \hline
14L & \begin{tabular}[c]{@{}c@{}}Subsegmetnal\\   (left)\end{tabular} & 14R & \begin{tabular}[c]{@{}c@{}}Subsegmetnal\\   (right)\end{tabular} \\ \hline
\multicolumn{4}{|c|}{\textbf{Mediastinum}} \\ \hline
\multicolumn{1}{|c|}{\textbf{Node}} & \multicolumn{3}{c|}{\textbf{Site}} \\ \hline
\multicolumn{1}{|c|}{7} & \multicolumn{3}{c|}{Subcarinal} \\ \hline
\multicolumn{1}{|c|}{8} & \multicolumn{3}{c|}{Paraesophageal} \\ \hline
\multicolumn{1}{|c|}{9} & \multicolumn{3}{c|}{Pulmonary Ligament} \\ \hline
\end{tabular}
\end{table}
According to the data acquisition protocol, the 21$\times$936 metastasis matrix is then divided into two sub-matrices based on the location of cancer nodule: for patients with nodules found at left lobe, "left lung + mediastinum" matrix will be used which consists of all the node sites within the "left lung" column in Table 1, as well as mediastinum. For patients with nodules found at right lobe, "right lung + mediastinum" matrix will be used in the similar way. 
The left and right metastasis matrices are then separately fed into the data summarization model as described in Section 3, resulting in two state matrices describing all the possible combination of the metastasis across lymph nodes. Finally, lung cancer metastasis pathways are inferred from the two summarization matrices, based on the graph inference method introduced in section 3. The result graphs of the left and right lung are visualized in ~\ref{fig:Fig5}. To validate the result graphs, we have conducted cross-validation experiments by randomly sampling 85$\%$ of the 936 patients for 100 times. These 100 sets of 800-patients data are then used to estimate the summarization matrices and to infer the network graphs. Cross-validation results show that edges inferred from the whole population can be consistently recovered from over 90$\%$ of the sub-sampled population: that is, every edge in the final pathway as shown in ~\ref{fig:Fig5} can be found in the graphs inferred from over 90 sets of the 800-patients data.
\clearpage
For comparison, we also visualize the results from GES on the same dataset by replacing missing values with zeros, as shown in ~\ref{fig:Fig6}. It can be seen that little correspondence could be found in the pathways inferred from left and right lobe by GES. More crucially, most of the intra-lobe connections are missing in the results from GES, indicating that it is incapable of recovering most edges from this real dataset. This observation is in accordance with the comparison of methods in the previous section of simulation study, where the false negative rate of GES is much higher than our proposed PathInf method.
\section{CONCLUSIONS}
In this work we develop a data summarization and graph inference model for learning underlying transition paths (i.e. network) among variables during a spreading process, addressing the challenge of massive incomplete data and lack of observation samples. The proposed PathInf model is tested and compared with GES on both simulation data, which shows superior performance (much lower false negative rate). PathInf is further applied on the lymph nodes sampling/dissection data from 936 patients, to establish an atlas for analyzing the lymphatic metastasis pathway among the node sites in left and right lung lobes. The pathways reveals novel discoveries for the lung clinical studies and can provide data-driven evidence for the empirical knowledge from the physicians. We are continuously working on the validation of the current discoveries from both the data science and clinical science perspective to establish an improved version of the current protocol for lung cancer examination and surgery. MATLAB code of the PathInf method is publicly available at GitHub.\\
One of the main limitation of the proposed PathInf method is that it cannot provide direction information on the pathways it inferred. In other words, it only infers correlation from co-existence of the positive variables but not causality between these variables. From a statistics perspective, it is impossible to infer causality from the given dataset as there are no temporal information involved: every sample is a "snapshot" of the patients condition during the transition process yet it is impossible to know when the snapshot is taken. Bayesian network models such as GES \cite{RN969} or PC \cite{doi:10.1177/089443939100900106} can infer the direction of edges yet they rely on the DAG assumption to do the inference. Works in \cite{Gomez-Rodriguez:2012:IND:2086737.2086741} which takes time-stamp of the events as input can infer the direction of pathways, but this type of time-stamp data is beyond our capability to acquire as it needs large of number patients to be examined more than once. Nevertheless, while the current results from the PathInf method are promising with novel discoveries, it would be important to infer the directed network from the data, as that is needed to determine the precise timing/sequence of the progression and to perform the simulation study of the metastasis (e.g. using Markov Chain). Thus, beyond this preliminary study, in the next step we are aiming to develop a more detailed inference modeling for the pathway directions based on more features from the patients which we already acquired but not yet been used, including tumor size and details of the pathologic analysis at each lymph site.
\begin{acks}
This work is supported by the Beijing Municipal Administration of Hospitals Clinical Medicine Development Special Funding\\
(ZYLX201509). In addition to this currently deployed project, the funding also covers the future data acquisition and analysis on a larger patient population in the next stage of study.
\end{acks}

\bibliographystyle{ACM-Reference-Format}
\bibliography{reference.bib}

\end{document}